\documentclass{article}

\usepackage{times}
\usepackage{latexsym}
\usepackage{microtype}
\usepackage{amsfonts}
\usepackage{amsthm}
\usepackage{amsmath}
\usepackage{graphicx}
\usepackage{subfigure}
\usepackage{booktabs}
\usepackage{multirow}
\usepackage{url}
\usepackage{hyperref}

\usepackage[accepted]{icml2021}
\icmltitlerunning{Grad-TTS: A Diffusion Probabilistic Model for Text-to-Speech}

\begin{document}

\twocolumn[
\icmltitle{Grad-TTS: A Diffusion Probabilistic Model for Text-to-Speech}

\icmlsetsymbol{equal}{*}

\begin{icmlauthorlist}
\icmlauthor{Vadim Popov}{equal,hw}
\icmlauthor{Ivan Vovk}{equal,hw,hse}
\icmlauthor{Vladimir Gogoryan}{hw,hse}
\icmlauthor{Tasnima Sadekova}{hw}
\icmlauthor{Mikhail Kudinov}{hw}
\end{icmlauthorlist}

\icmlaffiliation{hw}{Huawei Noah's Ark Lab, Moscow, Russia}
\icmlaffiliation{hse}{Higher School of Economics, Moscow, Russia}

\icmlcorrespondingauthor{Vadim Popov}{vadim.popov@huawei.com}
\icmlcorrespondingauthor{Ivan Vovk}{vovk.ivan@huawei.com}

\icmlkeywords{Text-to-Speech, score matching, diffusion probabilistic modelling, SDE}

\vskip 0.3in
]

\printAffiliationsAndNotice{\icmlEqualContribution}

\begin{abstract}
Recently, denoising diffusion probabilistic models and generative score matching have shown high potential in modelling complex data distributions while stochastic calculus has provided a unified point of view on these techniques allowing for flexible inference schemes. In this paper we introduce Grad-TTS, a novel text-to-speech model with score-based decoder producing mel-spectrograms by gradually transforming noise predicted by encoder and aligned with text input by means of Monotonic Alignment Search. The framework of stochastic differential equations helps us to generalize conventional diffusion probabilistic models to the case of reconstructing data from noise with different parameters and allows to make this reconstruction flexible by explicitly controlling trade-off between sound quality and inference speed. Subjective human evaluation shows that Grad-TTS is competitive with state-of-the-art text-to-speech approaches in terms of Mean Opinion Score. The code is publicly available at \url{https://github.com/huawei-noah/Speech-Backbones/tree/main/Grad-TTS}.

\end{abstract}

\section{Introduction}
\label{sec:intro}

Deep generative modelling proved to be effective in various machine learning fields, and speech synthesis is no exception. Modern text-to-speech (TTS) systems often consist of two parts designed as deep neural networks: the first part converts the input text into time-frequency domain acoustic features (\textit{feature generator}), and the second one synthesizes raw waveform conditioned on these features (\textit{vocoder}). Introduction of the conventional state-of-the-art autoregressive models such as Tacotron2 \cite{Tacotron2} used for feature generation and WaveNet \cite{WaveNet} used as vocoder marked the beginning of the neural TTS era. Later, other popular generative modelling frameworks such as Generative Adversarial Networks \cite{GANPaper} and Normalizing Flows \cite{NFlows} were used in the design of TTS engines for a parallel generation with comparable quality of the synthesized speech. 

Since the publication of the WaveNet paper \yrcite{WaveNet}, there have been various attempts to propose a parallel non-autoregressive vocoder, which could synthesize high-quality speech. Popular architectures based on Normalizing Flows like Parallel WaveNet \cite{ParallelWaveNet} and WaveGlow \cite{WaveGlow} managed to accelerate inference while keeping synthesis quality at a very high level but demonstrated fast synthesis on GPU devices only. Eventually, parallel GAN-based vocoders such as Parallel WaveGAN \cite{ParallelWaveGAN}, MelGAN \cite{MelGan}, and HiFi-GAN \cite{HiFi-GAN} greatly improved the performance of waveform generation on CPU devices. Furthermore, the latter model is reported to produce speech samples of state-of-the-art quality outperforming WaveNet.

Among feature generators, Tacotron2 \cite{Tacotron2} and Transformer-TTS \cite{TransformerTTS} enabled highly natural speech synthesis. Producing acoustic features frame by frame, they achieve almost perfect mel-spectrogram reconstruction from input text. Nonetheless, 
they often suffer from computational inefficiency and pronunciation issues coming from attention failures.
Addressing these problems, such models as FastSpeech \cite{FastSpeech} and Parallel Tacotron \cite{ParallelTacotron} substantially improved inference speed and pronunciation robustness by utilizing non-autoregressive architectures and building hard monotonic alignments from estimated token lengths. However, in order to learn character duration, they still require pre-computed alignment from the teacher model. Finally, the recently proposed Non-Attentive Tacotron framework \cite{NonAttentiveTacotron} managed to learn durations implicitly by employing the Variational Autoencoder concept.

Glow-TTS feature generator \cite{GlowTTS} based on Normalizing Flows can be considered as one of the most successful attempts to overcome pronunciation and computational latency issues typical for autoregressive solutions. Glow-TTS model made use of Monotonic Alignment Search algorithm (an adoption of Viterbi training \cite{Viterbi} finding the most likely hidden alignment between two sequences) proposed to map the input text to mel-spectrograms efficiently. The alignment learned by Glow-TTS is intentionally designed to avoid some of the pronunciation problems models like Tacotron2 suffer from. Also, in order to enable parallel synthesis, Glow-TTS borrows encoder architecture from Transformer-TTS \cite{TransformerTTS} and decoder architecture from Glow \cite{GlowNF}. Thus, compared with Tacotron2, Glow-TTS achieves much faster inference making fewer alignment mistakes. Besides, in contrast to other parallel TTS solutions such as FastSpeech, Glow-TTS does not require an external aligner to obtain token duration information as Monotonic Alignment Search (MAS) operates in an unsupervised way.

Lately, another family of generative models called Diffusion Probabilistic Models (DPMs) \cite{DiffusionBasic} has started to prove its capability to model complex data distributions such as images \cite{DDPM}, shapes \cite{Shapes}, graphs \cite{Graphs}, handwriting \cite{Handwriting}. 
The basic idea behind DPMs is as follows: we build a forward diffusion process by iteratively destroying original data until we get some simple distribution (usually standard normal), and then we try to build a reverse diffusion parameterized with a neural network so that it follows the trajectories of the reverse-time forward diffusion. Stochastic calculus offers a continuous easy-to-use framework for training DPMs \cite{SDE-main} and, which is perhaps more important, provides a number of flexible inference schemes based on numerical differential equation solvers.

As far as text-to-speech applications are concerned, two vocoders representing the DPM family showed impressive results in raw waveform reconstruction: WaveGrad \cite{WaveGrad} and DiffWave \cite{DiffWave} were shown to reproduce the fine-grained structure of human speech and match strong autoregressive baselines such as WaveNet in terms of synthesis quality while at the same time requiring much fewer sequential operations. However, despite such a success in neural vocoding, no feature generator based on diffusion probabilistic modelling is known so far. 

This paper introduces Grad-TTS, an acoustic feature generator with a score-based decoder using recent diffusion probabilistic modelling insights. In Grad-TTS, MAS-aligned encoder outputs are passed to the decoder that transforms Gaussian noise parameterized by these outputs into a mel-spectrogram. To cope with the task of reconstructing data from Gaussian noise with varying parameters, we write down a generalized version of conventional forward and reverse diffusions. One of the remarkable features of our model is that it provides explicit control of the trade-off between output mel-spectrogram quality and inference speed. In particular, we find that Grad-TTS is capable of generating mel-spectrograms of high quality with only as few as ten iterations of reverse diffusion, which makes it possible to outperform Tacotron2 in terms of speed on GPU devices. Additionally, we show that it is possible to train Grad-TTS as an end-to-end TTS pipeline (i.e., vocoder and feature generator are combined in a single model) by replacing its output domain from mel-spectrogram to raw waveform.

\section{Diffusion probabilistic modelling}
\label{sec:diffusion}

Loosely speaking, a process of the diffusion type is a stochastic process that satisfies a stochastic differential equation (SDE)

\begin{equation}
\label{eq:diff}
    dX_{t} = b(X_{t}, t)dt + a(X_{t}, t)dW_{t},
\end{equation}

where $W_{t}$ is the standard Brownian motion, $t\in [0, T]$ for some finite time horizon $T$, and coefficients $b$ and $a$ (called \textit{drift} and \textit{diffusion} correspondingly) satisfy certain measurability conditions. A rigorous definition of the diffusion type processes, as well as other notions from stochastic calculus we use in this section, can be found in \cite{SDE-book}.

\begin{figure}[ht]
\vskip 0.2in
\begin{center}
\centerline{\includegraphics[scale=0.6]{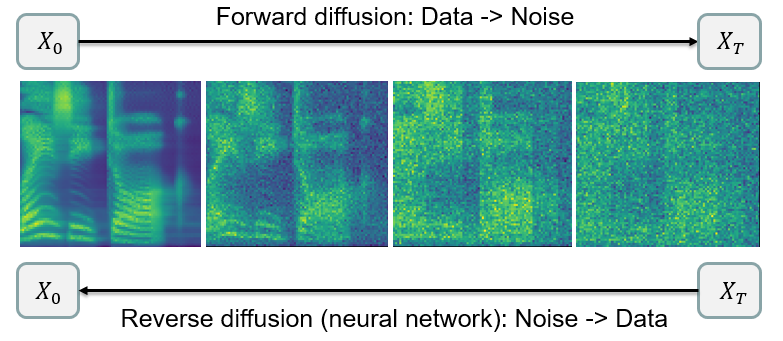}}
\caption{Diffusion probabilistic modelling for mel-spectrograms.}
\label{fig:diffusion}
\end{center}
\vskip -0.2in
\end{figure}

It is easy to find such a stochastic process that terminal distribution $Law(X_{T})$ converges to standard normal $\mathcal{N}(0,I)$ when $T\to \infty$ for any initial data distribution $Law(X_{0})$ ($I$ is $n\times n$ identity matrix and $n$ is data dimensionality). In fact, there are lots of such processes as it follows from the formulae given later in this section. Any process of the diffusion type with such property is called \textit{forward diffusion} and the goal of diffusion probabilistic modelling is to find a \textit{reverse diffusion} such that its trajectories closely follow those of the forward diffusion but in reverse time order. This is, of course, a much harder task than making Gaussian noise out of data, but in many cases it still can be accomplished if we parameterize reverse diffusion with a proper neural network. In this case, generation boils down to sampling random noise from $\mathcal{N}(0,I)$ and then just solving the SDE describing dynamics of the reverse diffusion with any numerical solver (usually a simple first-order Euler-Maruyama scheme \cite{SDE-numerical} is used). If forward and reverse diffusion processes have close trajectories, then the distribution of resulting samples will be very close to that of the data $Law(X_{0})$. This approach to generative modelling is summarized in Figure~\ref{fig:diffusion}.

Until recently, score-based and denoising diffusion probabilistic models were formalized in terms of Markov chains \cite{DiffusionBasic, ScoreBasedGeneration, DDPM, ScoreBasedImproved}. A unified approach introduced by Song et al. \yrcite{SDE-main} has demonstrated that these Markov chains actually approximated trajectories of stochastic processes satisfying certain SDEs. In our work, we follow this paper and define our DPM in terms of SDEs rather than Markov chains. As one can see later in Section \ref{sec:grad-tts}, the task we are solving suggests generalizing DPMs described in \cite{SDE-main} in such a way that for infinite time horizon forward diffusion transforms any data distribution into $\mathcal{N}(\mu,\Sigma)$ instead of $\mathcal{N}(0,I)$ for any given mean $\mu$ and diagonal covariance matrix $\Sigma$. So, the rest of this section contains the detailed description of the generalized forward and reverse diffusions we utilize as well as the loss function we optimize to train the reverse diffusion. All corresponding derivations can be found in \hyperref[sec:appendix]{Appendix}.

\subsection{Forward diffusion}
\label{subsec:fwd_diffusion}

First, we need to define a forward diffusion process that transforms any data into Gaussian noise given infinite time horizon $T$. If $n$-dimensional stochastic process $X_{t}$ satisfies the following SDE:

\begin{equation}
\label{eq:fwd_diffusion}
    dX_{t} = \frac{1}{2}\Sigma^{-1}(\mu - X_{t})\beta_{t}dt + \sqrt{\beta_{t}}dW_{t}, \ \ \ t\in [0,T]
\end{equation}

for non-negative function $\beta_{t}$, which we will refer to as noise schedule, vector $\mu$, and diagonal matrix $\Sigma$ with positive elements, then its solution (if it exists) is given by 

\begin{equation}
\begin{split}
\label{eq:solution}
    X_{t} &= \left(I - e^{-\frac{1}{2}\Sigma^{-1}\int_{0}^{t}\beta_{s}ds}\right)\mu
    + e^{-\frac{1}{2}\Sigma^{-1}\int_{0}^{t}\beta_{s}ds}X_{0}
    \\&+ \int_{0}^{t}{\sqrt{\beta_{s}}e^{-\frac{1}{2}\Sigma^{-1}\int_{s}^{t}{\beta_{u}du}}dW_{s}}.
\end{split}
\end{equation}

Note that the exponential of a diagonal matrix is just an element-wise exponential. Let

\begin{equation}
\begin{split}
\label{eq:mean}
    \rho(X_{0}, \Sigma, \mu, t) &= \left(I - e^{-\frac{1}{2}\Sigma^{-1}\int_{0}^{t}\beta_{s}ds}\right)\mu
    \\&+ e^{-\frac{1}{2}\Sigma^{-1}\int_{0}^{t}\beta_{s}ds}X_{0}
\end{split}
\end{equation}

and 

\begin{equation}
\label{eq:variance}
    \lambda(\Sigma, t) = \Sigma\left(I - e^{-\Sigma^{-1}\int_{0}^{t}\beta_{s}ds}\right).
\end{equation}

By properties of It\^o's integral conditional distribution of $X_{t}$ given $X_{0}$ is Gaussian:

\begin{equation}
\label{eq:sol_distribution}
    Law(X_{t}|X_{0}) = \mathcal{N}(\rho(X_{0}, \Sigma, \mu, t), \lambda(\Sigma, t)).
\end{equation}

It means that if we consider infinite time horizon then for \textit{any} noise schedule $\beta_{t}$ such that $\lim_{t\to \infty}e^{-\int_{0}^{t}\beta_{s}ds} = 0$ we have

\begin{equation}
\label{eq:sol_convergence}
    X_{t}|X_{0} \xrightarrow{d} \mathcal{N}(\mu, \Sigma).
\end{equation}

So, random variable $X_{t}$ converges in distribution to $\mathcal{N}(\mu, \Sigma)$ independently of $X_{0}$, and it is exactly the property we need: forward diffusion satisfying SDE (\ref{eq:fwd_diffusion}) transforms any data distribution $Law(X_{0})$ into Gaussian noise  $\mathcal{N}(\mu, \Sigma)$.

\subsection{Reverse diffusion}
\label{subsec:rev_diffusion}

While in earlier works on DPMs reverse diffusion was trained to approximate the trajectories of forward diffusion, Song et al. \yrcite{SDE-main} proposed to use the result by Anderson \yrcite{SDE-reverse}, who derived an explicit formula for reverse-time dynamics of a wide class of stochastic processes of the diffusion type. In our case, this result leads to the following SDE for the reverse diffusion:

\begin{equation}
\begin{split}
\label{eq:bwd_diffusion_sde}
    dX_{t} = & \left(\frac{1}{2}\Sigma^{-1}(\mu - X_{t}) - \nabla\log{p_{t}(X_{t})}\right)\beta_{t}dt
    \\&+ \sqrt{\beta_{t}}d\widetilde{W}_{t}, \qquad \qquad \qquad \qquad t\in [0,T],
\end{split}
\end{equation}

where $\widetilde{W}_{t}$ is the reverse-time Brownian motion and $p_{t}$ is the probability density function of random variable $X_{t}$. This SDE is to be solved backwards starting from terminal condition $X_{T}$.

Moreover, Song et al. \yrcite{SDE-main} have shown that instead of SDE (\ref{eq:bwd_diffusion_sde}), we can consider an ordinary differential equation

\begin{equation}
\label{eq:bwd_diffusion_ode}
    dX_{t} = \frac{1}{2}\left(\Sigma^{-1}(\mu - X_{t}) - \nabla\log{p_{t}(X_{t})}\right)\beta_{t}dt.
\end{equation}

Forward Kolmogorov equations corresponding to (\ref{eq:fwd_diffusion}) and (\ref{eq:bwd_diffusion_ode}) are identical, which means that the evolution of probability density functions of stochastic processes given by (\ref{eq:fwd_diffusion}) and (\ref{eq:bwd_diffusion_ode}) is the same.

Thus, if we have a neural network $s_{\theta}(X_{t}, t)$ that estimates the gradient of the log-density of noisy data $\nabla\log{p_{t}(X_{t})}$, then we can model data distribution $Law(X_{0})$ by sampling $X_{T}$ from $\mathcal{N}(\mu,\Sigma)$ and numerically solving either (\ref{eq:bwd_diffusion_sde}) or (\ref{eq:bwd_diffusion_ode}) backwards in time.

\subsection{Loss function}
\label{subsec:loss}

Estimating gradients of log-density of noisy data $X_{t}$ is often referred to as \textit{score matching}, and in recent papers \cite{ScoreBasedGeneration, ScoreBasedImproved} $L_{2}$ loss was used to approximate these gradients with a neural network. So, in our paper, we use the same type of loss.

\begin{figure*}[ht]
\begin{center}
\center{\includegraphics[width=1.0\linewidth]{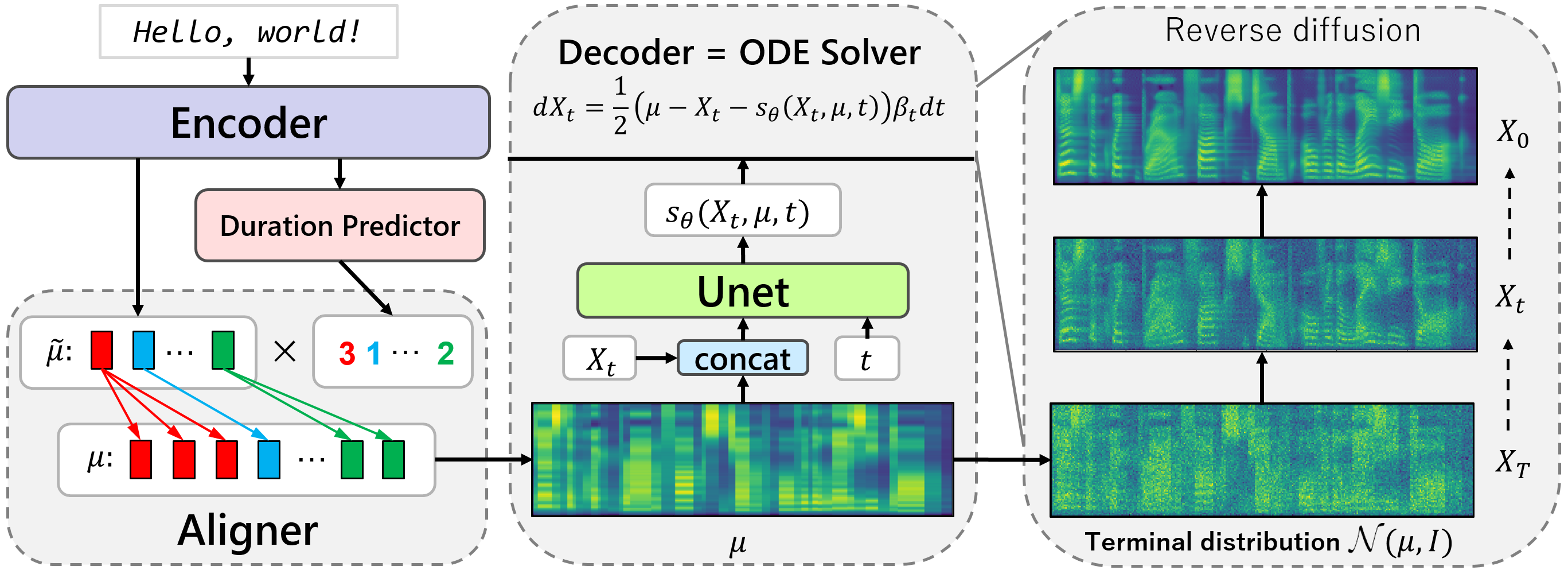}}
\end{center}
\caption{Grad-TTS inference scheme.}
\label{fig:main}
\end{figure*}

Due to the formula (\ref{eq:sol_distribution}), we can sample noisy data $X_{t}$ given only initial data $X_{0}$ without sampling intermediate values $\{X_{s}\}_{s<t}$. Moreover, $Law(X_{t}|X_{0})$ is Gaussian, which means that its log-density has a very simple closed form. If we sample $\epsilon_{t}$ from $\mathcal{N}(0,\lambda(\Sigma, t))$ and then put 

\begin{equation}
\label{eq:x_t}
    X_{t} = \rho(X_{0}, \Sigma, \mu, t) + \epsilon_t
\end{equation}

in accordance with (\ref{eq:sol_distribution}), then the gradient of log-density of noisy data in this point $X_{t}$ is given by

\begin{equation}
\label{eq:log_density}
    \nabla\log{p_{0t}(X_{t}|X_{0})} = -\lambda(\Sigma, t)^{-1}\epsilon_{t},
\end{equation}

where $p_{0t}(\cdot|X_{0})$ is the probability density function of the conditional distribution (\ref{eq:sol_distribution}). Thus, loss function corresponding to estimating the gradient of log-density of data $X_{0}$ corrupted with noise accumulated by time $t$ is

\begin{equation}
\label{eq:loss_t}
    \mathcal{L}_{t}(X_{0}) = \mathbb{E}_{\epsilon_{t}}\left[\left\Vert{s_{\theta}(X_{t},t) + \lambda(\Sigma, t)^{-1}\epsilon_{t}}\right\Vert_{2}^{2}\right],
\end{equation}

where $\epsilon_{t}$ is sampled from $\mathcal{N}(0,\lambda(\Sigma, t))$ and $X_{t}$ is calculated by formula (\ref{eq:x_t}).

\section{Grad-TTS}
\label{sec:grad-tts}

The acoustic feature generator we propose consists of three modules: encoder, duration predictor, and decoder. In this section, we will describe their architectures as well as training and inference procedures. The general approach is illustrated in Figure~\ref{fig:main}. Grad-TTS has very much in common with Glow-TTS \cite{GlowTTS}, a feature generator based on Normalizing Flows. The key difference lies in the principles the decoder relies on.

\subsection{Inference}
\label{subsec:inference}

An input text sequence $x_{1:L}$ of length $L$ typically consists of characters or phonemes, and we aim at generating mel-spectrogram $y_{1:F}$ where $F$ is the number of acoustic frames. In Grad-TTS, the encoder converts an input text sequence $x_{1:L}$ into a sequence of features $\tilde{\mu}_{1:L}$ used by the duration predictor to produce hard monotonic alignment $A$ between encoded text sequence $\tilde{\mu}_{1:L}$ and frame-wise features $\mu_{1:F}$. The function $A$ is a monotonic surjective mapping between $[1,F]\cap \mathbb{N}$ and $[1,L]\cap \mathbb{N}$, and we put $\mu_{j} = \tilde{\mu}_{A(j)}$ for any integer $j\in [1,F]$. Informally speaking, the duration predictor tells us how many frames each element of text input lasts. Monotonicity and surjectiveness of $A$ guarantee that the text is pronounced in the correct order without skipping any text input. As in all TTS models with duration predictor, it is possible to control synthesized speech tempo by multiplying predicted durations by some factor.

The output sequence $\mu = \mu_{1:F}$ is then passed to the decoder, which is a Diffusion Probabilistic Model. A neural network $s_{\theta}(X_{t}, \mu, t)$ with parameters $\theta$ defines an ordinary differential equation (ODE)

\begin{equation}
\label{eq:decoder}
    dX_{t} = \frac{1}{2}(\mu - X_{t} - s_{\theta}(X_{t}, \mu, t))\beta_{t}dt,
\end{equation}

which is solved backwards in time using the first-order Euler scheme. The sequence $\mu$ is also used to define the terminal condition $X_{T}\sim \mathcal{N}(\mu, I)$. Noise schedule $\beta_{t}$ and time horizon $T$ are some pre-defined hyperparameters whose choice mostly depends on the data, while step size $h$ in the Euler scheme is a hyperparameter that can be chosen after Grad-TTS is trained. It expresses the trade-off between the quality of output mel-spectrograms and inference speed.

Reverse diffusion in Grad-TTS evolves according to equation (\ref{eq:decoder}) for the following reasons:

\begin{itemize}
    \item We obtained better results in practice when using dynamics (\ref{eq:bwd_diffusion_ode}) instead of (\ref{eq:bwd_diffusion_sde}): for small values of step size $h$, they performed equally well, while for larger values the former led to much better sounding results.
    \item We chose $\Sigma = I$ to simplify the whole feature generation pipeline.
    \item We used $\mu$ as an additional input to the neural network $s_{\theta}(X_{t}, \mu, t)$. It follows from (\ref{eq:log_density}) that the neural network $s_{\theta}$ essentially tries to predict Gaussian noise added to data $X_{0}$ observing only noisy data $X_{t}$. So, if for every time $t$ we supply $s_{\theta}$ with an additional knowledge of how the limiting noise $\lim_{T\to \infty} Law(X_{T}|X_{0})$ looks like (note that it is different for different text input), then this network can make more accurate predictions of noise at time $t\in [0,T]$.
\end{itemize}

We also found it beneficial for the model performance to introduce a temperature hyperparameter $\tau$ and to sample terminal condition $X_{T}$ from $\mathcal{N}(\mu,\tau^{-1}I)$ instead of $\mathcal{N}(\mu,I)$. Tuning $\tau$ can help to keep the quality of output mel-spectrograms at the same level when using larger values of step size $h$.

\subsection{Training}
\label{subsec:training}

One of Grad-TTS training objectives is to minimize the distance between aligned encoder output $\mu$ and target mel-spectrogram $y$ because the inference scheme that has just been described suggests to start decoding from random noise $\mathcal{N}(\mu,I)$. Intuitively, it is clear that decoding is easier if we start from noise, which is already close to the target $y$ in some sense.

If aligned encoder output $\mu$ is considered to parameterize an input noise the decoder starts from, it is natural to regard encoder output $\tilde{\mu}$ as a normal distribution $\mathcal{N}(\tilde{\mu},I)$, which leads to a negative log-likelihood encoder loss:

\begin{equation}
\label{eq:loss_enc}
    \mathcal{L}_{enc} = -\sum_{j=1}^{F}{\log{\varphi(y_{j};\tilde{\mu}_{A(j)}, I)}},
\end{equation}

where $\varphi(\cdot;\tilde{\mu}_{i},I)$ is a probability density function of $\mathcal{N}(\tilde{\mu}_{i}, I)$. Although other types of losses are also possible, we have chosen $\mathcal{L}_{enc}$ (which actually reduces to Mean Square Error criterion) because of this probabilistic interpretation. In principle, it is even possible to train Grad-TTS without any encoder loss at all and let the diffusion loss described further do all the job of generating realistic mel-spectrograms, but in practice we observed that in the absence of $\mathcal{L}_{enc}$ Grad-TTS failed to learn alignment.

The encoder loss $\mathcal{L}_{enc}$ has to be optimized with respect to both encoder parameters and alignment function $A$. Since it is hard to do a joint optimization, we apply an iterative approach proposed by Kim et al. \yrcite{GlowTTS}. Each iteration of optimization consists of two steps: (i) searching for an optimal alignment $A^{*}$ given fixed encoder parameters; (ii) fixing this alignment $A^{*}$ and taking one step of stochastic gradient descent to optimize loss function with respect to encoder parameters. We use Monotonic Alignment Search at the first step of this approach. MAS utilizes the concept of dynamic programming to find an optimal (from the point of view of loss function $\mathcal{L}_{enc}$) monotonic surjective alignment. This algorithm is described in detail in \cite{GlowTTS}.

To estimate the optimal alignment $A^{*}$ at inference, Grad-TTS employs the duration predictor network. As in \cite{GlowTTS}, we train the duration predictor $DP$ with Mean Square Error (MSE) criterion in logarithmic domain:

\begin{equation}
\begin{split}
\label{eq:loss_dur}
    d_{i} = &\log{\sum_{j=1}^{F}{\mathbb{I}_{\{A^{*}(j)=i\}}}}, \ \ \ i=1,..,L,
    \\& \mathcal{L}_{dp} = MSE(DP(sg[\tilde{\mu}]), d),
\end{split}
\end{equation}

where $\mathbb{I}$ is an indicator function, $\tilde{\mu}=\tilde{\mu}_{1:L}$, $d=d_{1:L}$ and stop gradient operator $sg[\cdot]$ is applied to the inputs of the duration predictor to prevent $\mathcal{L}_{dp}$ from affecting encoder parameters.

As for the loss related to the DPM, it is calculated using formulae from Section \ref{sec:diffusion}. As already mentioned, we put $\Sigma = I$, so the distribution of noisy data (\ref{eq:sol_distribution}) simplifies, and its covariance matrix becomes just an identity matrix $I$ multiplied by a scalar

\begin{equation}
\label{eq:variance_simplified}
    \lambda_{t} = 1 - e^{-\int_{0}^{t}{\beta_{s}ds}}.
\end{equation}

The overall diffusion loss function $\mathcal{L}_{diff}$ is the expectation of weighted losses associated with estimating gradients of log-density of noisy data at different times $t\in [0,T]$:

\begin{equation}
\label{eq:loss_diff}
    \mathcal{L}_{diff} = \mathbb{E}_{X_{0}, t}\left[\lambda_{t}\mathbb{E}_{\xi_{t}}\left[\left\Vert s_{\theta}(X_{t}, \mu, t) + \frac{\xi_{t}}{\sqrt{\lambda_{t}}} \right\Vert_{2}^{2}\right]\right],
\end{equation}

where $X_{0}$ stands for target mel-spectrogram $y$ sampled from training data, $t$ is sampled from uniform distribution on $[0,T]$, $\xi_{t}$ -- from $\mathcal{N}(0,I)$ and the formula 

\begin{equation}
\label{eq:x_t_sampling}
X_{t}=\rho(X_{0},I,\mu,t) + \sqrt{\lambda_{t}}\xi_{t}
\end{equation}

is used to get noisy data $X_{t}$ according to the distribution (\ref{eq:sol_distribution}). The above formulae (\ref{eq:loss_diff}) and (\ref{eq:x_t_sampling}) follow from (\ref{eq:loss_t}) and (\ref{eq:x_t}) by substitution $\epsilon_{t}=\sqrt{\lambda_{t}}\xi_{t}$. We use losses (\ref{eq:loss_t}) with weights $\lambda_{t}$ according to the common heuristics that these weights should be proportional to $1/\mathbb{E}\left[\left\Vert\nabla\log{p_{0t}(X_{t}|X_{0})}\right\Vert_{2}^{2}\right]$.

To sum it up, the training procedure consists of the following steps:

\begin{itemize}
    \item Fix the encoder, duration predictor, and decoder parameters and run MAS algorithm to find the alignment $A^{*}$ that minimizes $\mathcal{L}_{enc}$.
    \item Fix the alignment $A^{*}$ and minimize $\mathcal{L}_{enc} + \mathcal{L}_{dp} + \mathcal{L}_{diff}$ with respect to encoder, duration predictor, and decoder parameters.
    \item Repeat the first two steps till convergence.
\end{itemize}

\subsection{Model architecture}
\label{subsection:architecture}

As for the encoder and duration predictor, we use exactly the same architectures as in Glow-TTS, which in its turn borrows the structure of these modules from Transformer-TTS \cite{TransformerTTS} and FastSpeech \cite{FastSpeech} correspondingly. The duration predictor consists of two convolutional layers followed by a projection layer that predicts the logarithm of duration. The encoder is composed of a pre-net, $6$ Transformer blocks with multi-head self-attention, and the final linear projection layer. The pre-net consists of $3$ layers of convolutions followed by a fully-connected layer.

The decoder network $s_{\theta}$ has the same U-Net architecture \cite{UNet} used by Ho et al. \yrcite{DDPM} to generate $32\times 32$ images, except that we use twice fewer channels and three feature map resolutions instead of four to reduce model size. In our experiments we use $80$-dimensional mel-spectrograms, so $s_{\theta}$ operates on resolutions $80\times F$, $40\times F/2$ and $20\times F/4$. We zero-pad mel-spectrograms if the number of frames $F$ is not a multiple of $4$. Aligned encoder output $\mu$ is concatenated with U-Net input $X_{t}$ as an additional channel.

\section{Experiments}
\label{sec:exp}

LJSpeech dataset \cite{LJSpeech} containing approximately $24$ hours of English female voice recordings sampled at $22.05$kHz was used to train the Grad-TTS model. The test set contained around $500$ short audio recordings (duration less than $10$ seconds each). The input text was phonemized before passing to the encoder; as for the output acoustic features, we used conventional $80$-dimensional mel-spectrograms. We tried training both on original and normalized mel-spectrograms and found that the former performed better. Grad-TTS was trained for $1.7m$ iterations on a single GPU (NVIDIA RTX $2080$ Ti with $11$GB memory) with mini-batch size $16$. We chose Adam optimizer and set the learning rate to $0.0001$.

\begin{figure}[ht]
\vskip 0.1in
\begin{center}
\centerline{\includegraphics[scale=0.33]{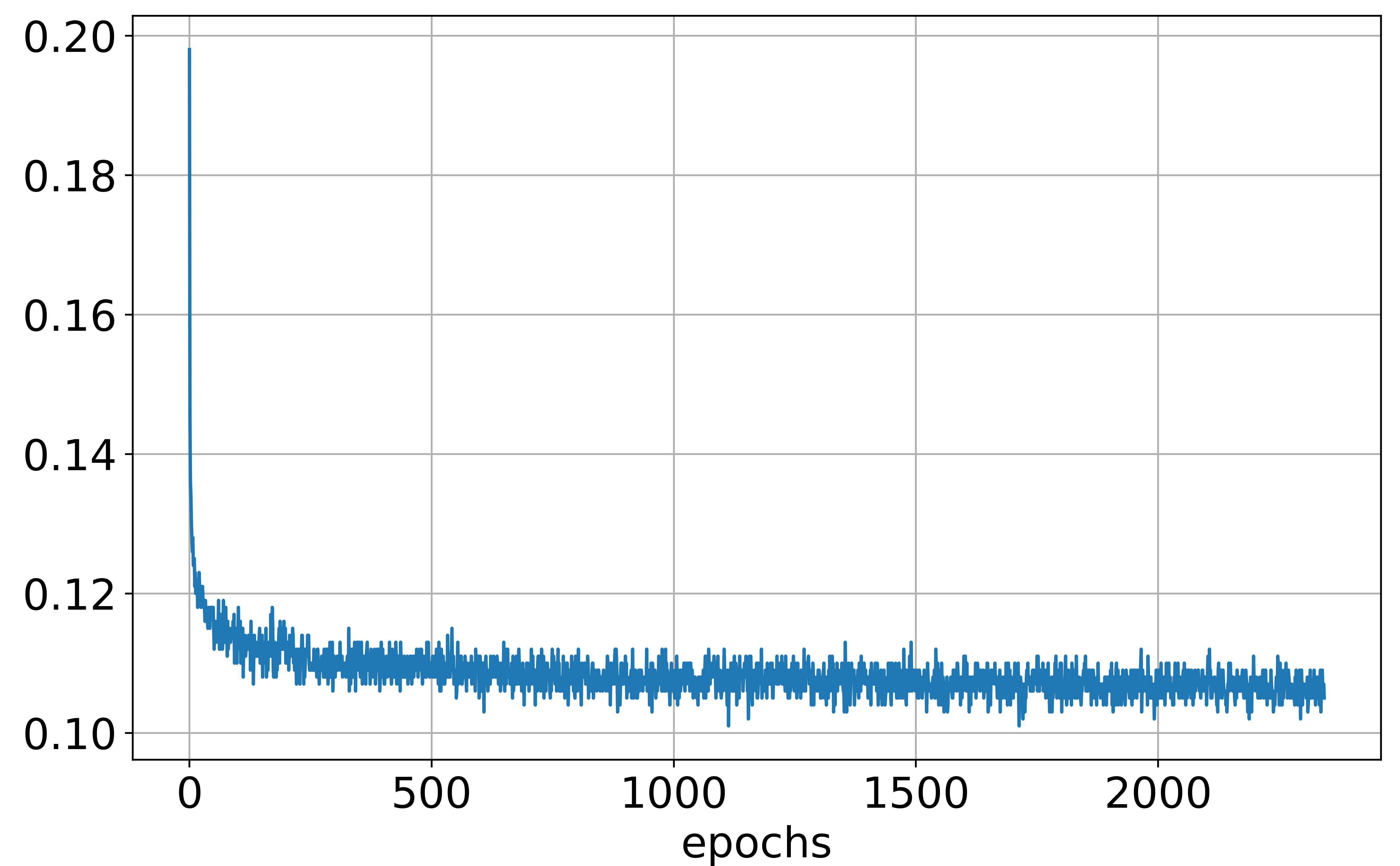}}
\caption{Diffusion loss at training.}
\label{fig:training}
\end{center}
\vskip -0.1in
\end{figure}

We would like to mention several important things about Grad-TTS training:

\begin{itemize}
    \item We chose $T=1$, $\beta_{t}=\beta_{0}+(\beta_{1}-\beta_{0})t$, $\beta_{0}=0.05$ and $\beta_{1}=20$.
    \item As in \cite{GAN-TTS1, GAN-TTS2}, we use random mel-spectrogram segments of fixed length ($2$ seconds in our case) as training targets $y$ to allow for memory-efficient training. However, MAS and the duration predictor still use whole mel-spectrograms.
    \item Although diffusion loss $\mathcal{L}_{diff}$ seems to converge very slowly after the beginning epochs, as shown on Figure~\ref{fig:training}, such long training is essential to get a good model because the neural network $s_{\theta}$ has to learn to estimate gradients accurately for all $t\in [0,1]$. Two models with almost equal diffusion losses can produce mel-spectrograms of very different quality: inaccurate predictions for a small subset $S\subset [0,1]$ may have a small impact on $\mathcal{L}_{diff}$ but be crucial for the output mel-spectrogram quality if ODE solver involves calculating $s_{\theta}$ in at least one point belonging to $S$.
\end{itemize}

Once trained, Grad-TTS enables the trade-off between quality and inference speed due to the ability to vary the number of steps $N$ the decoder takes to solve ODE (\ref{eq:decoder}) at inference. So, we evaluate four models which we denote by Grad-TTS-N where $N\in [4, 10, 100, 1000]$. We use $\tau=1.5$ at synthesis for all four models. As baselines, we take an official implementation of Glow-TTS \cite{GlowTTS}, the model which resembles ours to the most extent among the existing feature generators, FastSpeech \cite{FastSpeech}, and state-of-the-art Tacotron2 \cite{Tacotron2}. Recently proposed HiFi-GAN \cite{HiFi-GAN} is known to provide excellent sound quality, so we use this vocoder with all models we compare.

\subsection{Subjective evaluation}
\label{subsec:subj}
To make subjective evaluation of TTS models, we used the crowdsourcing platform Amazon Mechanical Turk. For Mean Opinion Score (MOS) estimation we synthesized $40$ sentences from the test set with each model. The assessors were asked to estimate the quality of synthesized speech on a nine-point Likert scale, the lowest and the highest scores being $1$ point (``Bad'') and $5$ points (``Excellent'') with a step of $0.5$ point. To ensure the reliability of the obtained results, only Master assessors were assigned to complete the listening test. Each audio was evaluated by $10$ assessors. A small subset of speech samples used in the test is available at \url{https://grad-tts.github.io/}.

\begin{table}[H]
\caption{Ablation study of proposed generalized diffusion framework. Grad-TTS reconstructing data from $\mathcal{N}(0, I)$ for $N$ reverse diffusion iterations is compared with the baseline Grad-TTS-10 -- the model reconstructing data from $\mathcal{N}(\mu, I)$ for $10$ iterations.}
\begin{center}
\begin{tabular}{|c|c|c|c|}
\hline
$N$ &Worse, \% &Identical, \% &Better, \% \\ \hline
$10$ &$93.8$ &$0.5$ &$5.7$ \\ \hline
$20$ &$82.3$ &$2.9$ &$14.8$ \\ \hline
$50$ &$60.3$ &$5.7$ &$34.0$ \\ \hline
\end{tabular}
\end{center}
\label{tab:preference_test}
\end{table}

MOS results with $95\%$ confidence intervals are presented in Table~\ref{tab:main}. It demonstrates that although the quality of the synthesized speech gets better when we use more iterations of the reverse diffusion, the quality gain becomes marginal starting from a certain number of iterations. In particular, there is almost no difference between Grad-TTS-1000 and Grad-TTS-10 in terms of MOS, while the gap between Grad-TTS-10 and Grad-TTS-4 ($4$ was the smallest number of iterations leading to satisfactory quality) is much more significant. As for other feature generators, Grad-TTS-10 is competitive with all compared models, including state-of-the-art Tacotron2. Furthermore, Grad-TTS-1000 achieves almost natural synthesis with MOS being less than that for ground truth recordings by only $0.1$. We would like to note that the relatively low results of FastSpeech could possibly be explained by the fact that we used its unofficial implementation \url{https://github.com/xcmyz/FastSpeech}.

\begin{table*}[ht]
\caption{Model comparison.}
\begin{center}
\begin{tabular}{|c|c|c|c|c|c|}
\hline
Model &Enc params\footnote{} &Dec params &RTF &Log-likelihood &MOS \\ \hline
Grad-TTS-1000 & \multirow{4}{*}{$7.2m$} & \multirow{4}{*}{$7.6m$} &$3.663$ & \multirow{4}{*}{$\mathbf{0.174\pm 0.001}$} &$\mathbf{4.44\pm 0.05}$ \\ \cline{1-1}\cline{4-4}\cline{6-6}
Grad-TTS-100 & & &$0.363$ & &$4.38\pm 0.06$ \\ \cline{1-1}\cline{4-4}\cline{6-6}
Grad-TTS-10 & & &$0.033$ & &$4.38\pm 0.06$ \\ \cline{1-1}\cline{4-4}\cline{6-6}
Grad-TTS-4 & & &$0.012$ & &$3.96\pm 0.07$ \\ \hline
Glow-TTS &$7.2m$ &$21.4m$ &$0.008$ &$0.082$ &$4.11\pm 0.07$ \\ \hline
FastSpeech & \multicolumn{2}{c}{$24.5m$} \vline &$\mathbf{0.004}$ &$-$ &$3.68\pm 0.09$ \\ \hline
Tacotron2 & \multicolumn{2}{c}{$28.2m$} \vline &$0.075$ &$-$ &$4.32\pm 0.07$ \\ \hline
Ground Truth & \multicolumn{2}{c}{$-$} \vline &$-$ &$-$ &$4.53\pm 0.06$ \\ \hline
\end{tabular}
\end{center}
\label{tab:main}
\end{table*}

To verify the benefits of the proposed generalized DPM framework we trained the model with the same architecture as Grad-TTS to reconstruct mel-spectrograms from $\mathcal{N} (0, I)$ instead of $\mathcal{N} (\mu, I)$. The preference test provided in Table \ref{tab:preference_test} shows that Grad-TTS-10 is significantly better ($p < 0.005$ in sign test) than this model taking $10$, $20$ and even $50$ iterations of the reverse diffusion. It demonstrates that the model trained to generate from $\mathcal{N} (0, I)$ needs more steps of ODE solver to get high-quality mel-spectrograms than Grad-TTS we propose. We believe this is because the task of reconstructing mel-spectrogram from pure noise $\mathcal{N} (0, I)$ is more difficult than the one of reconstructing it from its noisy copy $\mathcal{N} (\mu, I)$. One possible objection could be that the model trained with $\mathcal{N} (0, I)$ as terminal distribution can just add $\mu$ to this noise at the first step of sampling (it is possible because $s_{\theta}$ has $\mu$ as its input) and then repeat the same steps as our model to generate data from $N (\mu, I)$. In this case, it would generate mel-spectrograms of the same quality as our model taking only one step more. However, this argument is wrong, since reverse diffusion removes noise not arbitrarily, but according to the reverse trajectories of the forward diffusion. Since forward diffusion adds noise gradually, reverse diffusion has to remove noise gradually as well, and the first step of the reverse diffusion cannot be adding $\mu$ to Gaussian noise with zero mean because the last step of the forward diffusion is not a jump from $\mu$ to zero.

\begin{figure}[ht]
\vskip 0.1in
\begin{center}
\centerline{\includegraphics[scale=0.33]{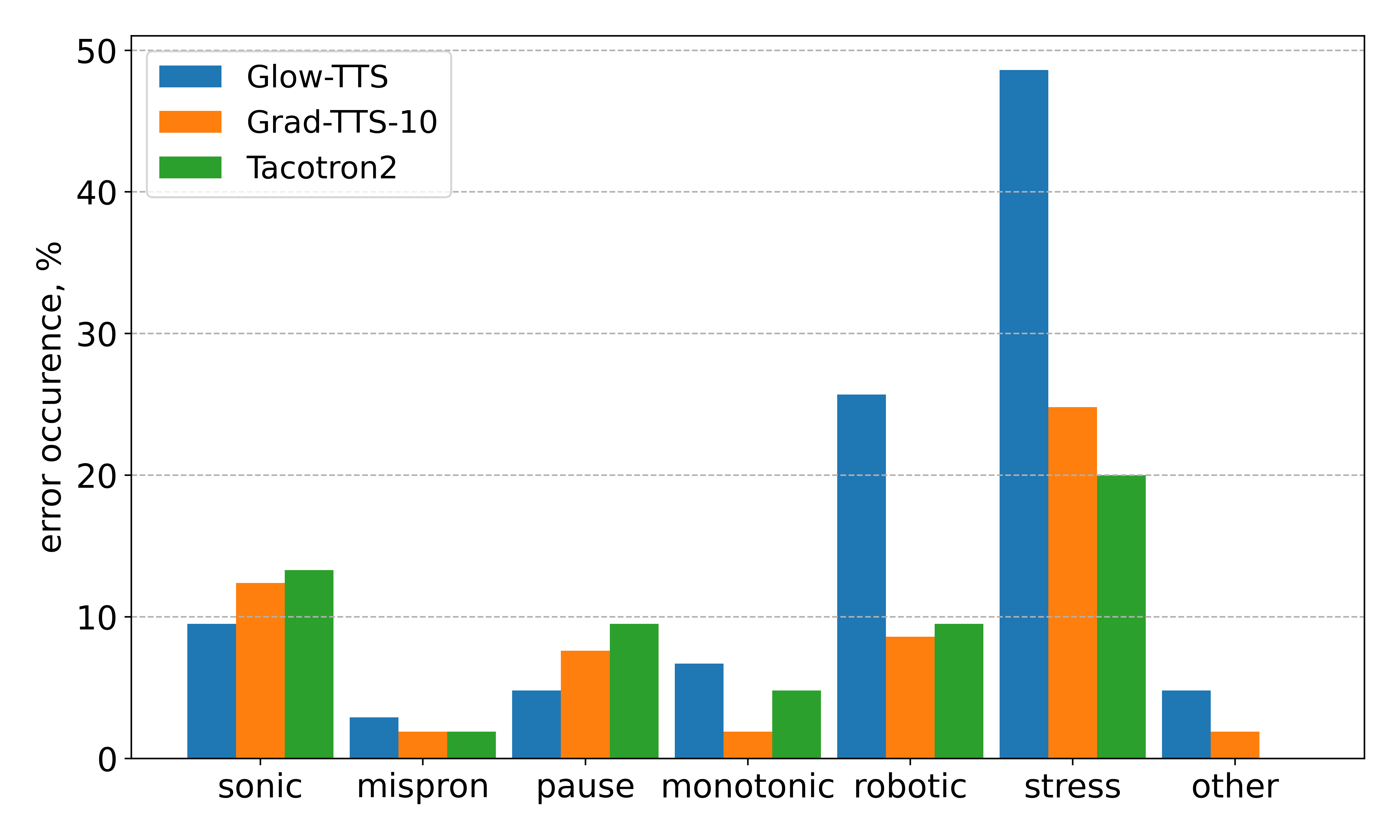}}
\caption{Typical errors occurrence.}
\label{fig:qualitative}
\end{center}
\vskip -0.1in
\end{figure}

We also made an attempt to estimate what kinds of mistakes are characteristic of certain models. We compared Tacotron2, Glow-TTS, and Grad-TTS-10 as the fastest version of our model with high synthesis quality. Each record was estimated by $5$ assessors. Figure \ref{fig:qualitative} demonstrates the results of the multiple-choice test whose participants had to choose which kinds of errors (if any) they could hear: sonic artifacts like clicking sounds or background noise (``sonic'' in the figure), mispronunciation of words/phonemes (``mispron''), unnatural pauses (``pause''), monotone speech (``monotonic''), robotic voice (``robotic''), wrong word stressing (``stress'') or others. It is clear from the figure that Glow-TTS frequently stresses words in a wrong way, and the sound it produces is perceived as ``robotic'' in around a quarter of cases. These are the major factors that make Glow-TTS performance inferior to that of Grad-TTS and Tacotron2, which in their turn have more or less the same drawbacks in terms of synthesis quality.

\subsection{Objective evaluation}
\label{subsec:obj}
Although DPMs can be shown to maximize weighted variational lower bound \cite{DDPM} on data log-likelihood, they do not explicitly optimize exact data likelihood. In spite of this, Song et al. \yrcite{SDE-main} show that it is still possible to calculate it using the instantaneous change of variables formula \cite{NODE} if we look at DPMs from the ``continuous'' point of view. However, it is necessary to use Hutchinson's trace estimator to make computations feasible, so in Table~\ref{tab:main} log-likelihood for Grad-TTS comes with a $95\%$ confidence interval.

We randomly chose $50$ sentences from the test set and calculated their average log-likelihood under two probabilistic models we consider -- Glow-TTS and Grad-TTS. Interestingly, Grad-TTS achieves better log-likelihood than Glow-TTS even though the latter has a decoder with $3$x larger capacity and was trained to maximize exact data likelihood. Similar phenomena were observed by Song et al. \yrcite{SDE-main} in the image generation task.

\subsection{Efficiency estimation}
\label{subsec:efficiency}

\begin{figure}[!ht]
\vskip 0.1in
\begin{center}
\centerline{\includegraphics[scale=0.34]{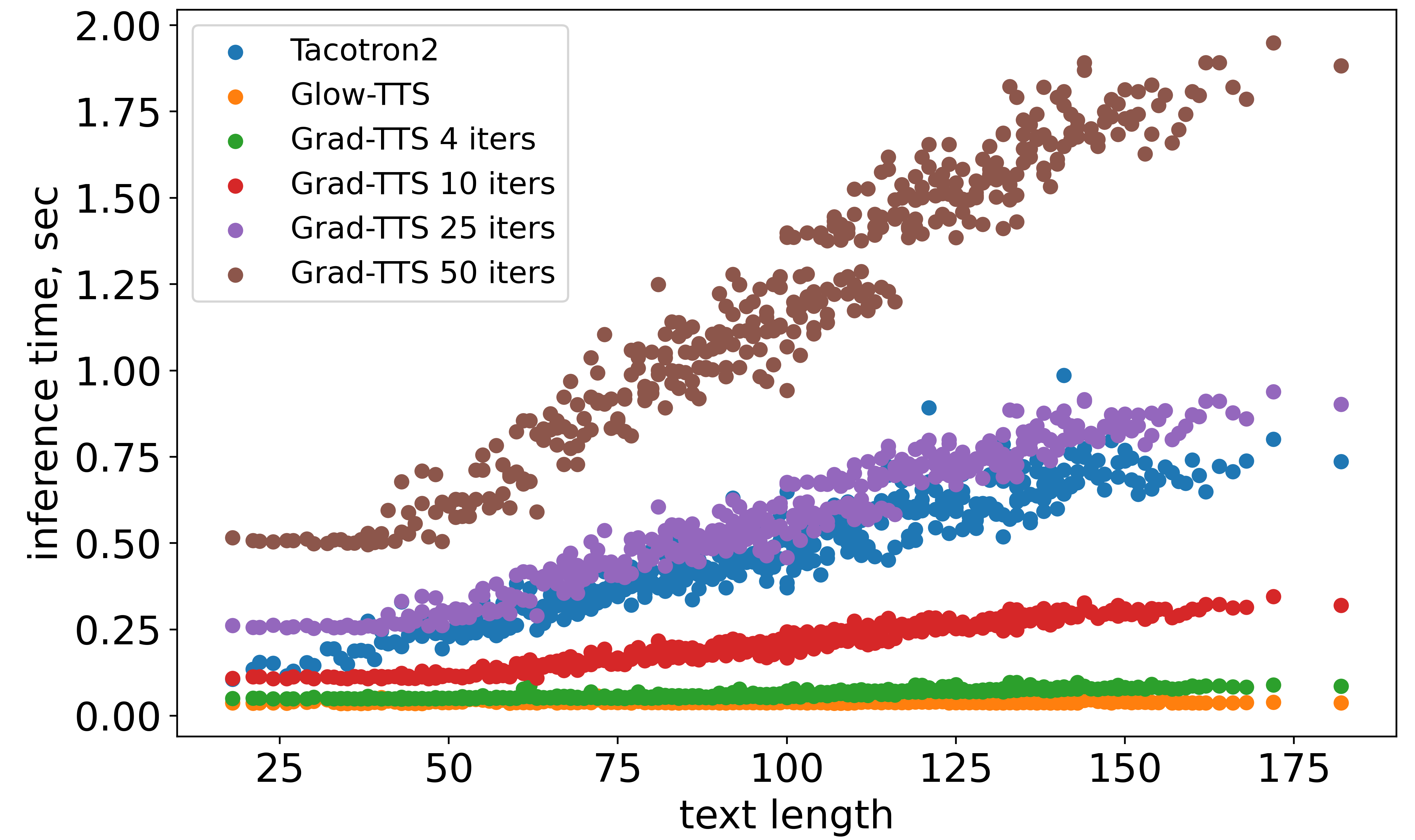}}
\caption{Inference speed comparison. Text length is given in characters.}
\label{fig:speed}
\end{center}
\vskip -0.1in
\end{figure}

We assess the efficiency of the proposed model in terms of Real-Time Factor (RTF is how many seconds it takes to generate one second of audio) computed on GPU and the number of parameters. Table~\ref{tab:main} contains efficiency information for all models under comparison.
Additional information regarding absolute inference speed dependency on the input text length is given in Figure~\ref{fig:speed}.

Due to its flexibility at inference, Grad-TTS is capable of real-time synthesis on GPU: if the number of decoder steps is less than $100$, it reaches RTF $<0.37$. Moreover, although it cannot compete with Glow-TTS and FastSpeech in terms of inference speed, it still can be approximately twice faster than Tacotron2 if we use $10$ decoder iterations sufficient for getting high-fidelity mel-spectrograms. Besides, Grad-TTS has around $15m$ parameters, thus being significantly smaller than other feature generators we compare.

\subsection{End-to-end TTS}
\label{subsec:e2e}

The results of our preliminary experiments show that it is also possible to train an end-to-end TTS model as a DPM. In brief, we moved from U-Net to WaveGrad \cite{WaveGrad} in Grad-TTS decoder: the overall architecture resembles WaveGrad conditioned on the aligned encoder output $\mu$ instead of ground truth mel-spectrograms $y$ as in original WaveGrad. Although synthesized speech quality is fair enough, it cannot compete with the results reported above, so we do not include our end-to-end model in the listening test but provide demo samples at \url{https://grad-tts.github.io/}.
\footnotetext{Encoder and duration predictor parameters are calculated together.}

\section{Future work}
\label{sec:future}

End-to-end speech synthesis results reported above show that it is a promising future research direction for text-to-speech applications. However, there is also much room for investigating general issues regarding DPMs.

In the analysis in Section \ref{sec:diffusion}, we always assume that both forward and reverse diffusion processes exist, i.e., SDEs (\ref{eq:fwd_diffusion}) and (\ref{eq:bwd_diffusion_sde}) have strong solutions. It applies some Lipschitz-type constraints \cite{SDE-book} on noise schedule $\beta_{t}$ and, what is more important, on the neural network $s_{\theta}$. Wasserstein GANs offer an encouraging example of incorporating Lipschitz constraints into neural networks training \cite{WGAN-GP}, suggesting that similar techniques may improve DPMs.

Little attention has been paid so far to the choice of the noise schedule $\beta_{t}$ -- most researchers use a simple linear schedule. Also, it is mostly unclear how to choose weights for losses (\ref{eq:loss_t}) at time $t$ in the global loss function optimally. A thorough investigation of such practical questions is crucial as it can facilitate applying DPMs to new machine learning problems.

\section{Conclusion}
\label{sec:conclusion}

We have presented Grad-TTS, the first acoustic feature generator utilizing the concept of diffusion probabilistic modelling. The main generative engine of Grad-TTS is the diffusion-based decoder that transforms Gaussian noise parameterized with the encoder output into mel-spectrogram while alignment is performed with Monotonic Alignment Search. The model we propose allows to vary the number of decoder steps at inference, thus providing a tool to control the trade-off between inference speed and synthesized speech quality. Despite its iterative decoding, Grad-TTS is capable of real-time synthesis. Moreover, it can generate mel-spectrograms twice faster than Tacotron2 while keeping synthesis quality competitive with common TTS baselines.

\bibliography{example_paper}
\bibliographystyle{icml2021}

\newpage
\onecolumn
\section*{Appendix}
\label{sec:appendix}
We include an appendix with detailed derivations, proofs and additional information. Our proposed diffusion probabilistic framework employs generalized terminal distribution $\mathcal{N} (\mu, \Sigma)$ instead of $\mathcal{N} (0, I)$ as proposed by Song et al. \yrcite{SDE-main}. The derivation for the solution (\ref{eq:solution}) of SDE (\ref{eq:fwd_diffusion}) that transforms the original data distribution to the terminal distribution is described in Appendix \ref{app:solving_forward_diff_sde}. In Appendix \ref{app:derivation_of_cond_dist} we also derive the distribution which the solution (\ref{eq:solution}) for the diffused data $X_t$ follows. Then, the goal of diffusion probabilistic modelling is to reconstruct the reverse-time trajectories of the forward diffusion process, and Song et al. \yrcite{SDE-main} showed that these dynamics can follow two different differential equations: either SDE (\ref{eq:bwd_diffusion_sde}) proposed by Anderson \yrcite{SDE-reverse} or ODE (\ref{eq:bwd_diffusion_ode}). So, Appendix \ref{app:reverse_dynamics} contains these differential equations for $\mathcal{N} (\mu, \Sigma)$ serving as terminal distribution. They depend on time-dependent gradient field $\nabla\log{p_{0t}(X_{t}|X_{0})}$ supposed to be modelled using neural network. In order to train it, we show how to compute the gradient in Appendix \ref{app:score_estimation}.

\renewcommand{\thesubsection}{\Alph{subsection}}

\subsection{Solving forward diffusion SDE}
\label{app:solving_forward_diff_sde}

Forward diffusion SDE is given by 

\begin{equation}
    dX_{t} = \frac{1}{2}\Sigma^{-1}(\mu - X_{t})\beta_{t}dt + \sqrt{\beta_{t}}dW_{t}, \ \ \ \ t\in [0,T],
\end{equation}

where $X_{t}$ is $n$-dimensional stochastic process, $W_{t}$ is the standard $n$-dimensional Brownian motion, $\mu = (\mu_{1} ... \mu_{n})^{\mathbf{T}}$ is $n$-dimensional vector, $\Sigma$ is $n\times n$ diagonal matrix with positive diagonal elements $\{\sigma^{2}_{ii}\}_{1}^{n}$ and noise schedule $\beta_{t}$ is non-negative function $[0,T]\rightarrow \mathbb{R}^{+}$. Consider change of variables $Y_{t} = X_{t} - \mu$. Then we can rewrite forward diffusion SDE as

\begin{equation}
    dY_{t} = -\frac{1}{2}\Sigma^{-1}Y_{t}\beta_{t}dt + \sqrt{\beta_{t}}dW_{t}.
\end{equation}

For every $i=1,..,n$ we have

\begin{equation}
\begin{split}
    d\left(e^{\frac{1}{2\sigma^{2}_{ii}}\int_{0}^{t}{\beta_{s}ds}}Y_{t}^{i}\right) &= e^{\frac{1}{2\sigma^{2}_{ii}}\int_{0}^{t}{\beta_{s}ds}}\cdot \frac{1}{2\sigma^{2}_{ii}}\beta_{t}Y_{t}^{i}dt + e^{\frac{1}{2\sigma^{2}_{ii}}\int_{0}^{t}{\beta_{s}ds}}\cdot \left(-\frac{1}{2\sigma^{2}_{ii}}Y_{t}^{i}\beta_{t}dt + \sqrt{\beta_{t}}dW_{t}^{i}\right) =\\&= e^{\frac{1}{2\sigma^{2}_{ii}}\int_{0}^{t}{\beta_{s}ds}}\sqrt{\beta_{t}}dW_{t}^{i}.
\end{split}
\end{equation}

Exponential of a diagonal matrix is just element-wise exponential, so we can rewrite it in multidimensional form as

\begin{equation}
    d\left(e^{\frac{1}{2}\Sigma^{-1}\int_{0}^{t}{\beta_{s}ds}}Y_{t}\right) = \sqrt{\beta_{t}}e^{\frac{1}{2}\Sigma^{-1}\int_{0}^{t}{\beta_{s}ds}}dW_{t} \implies e^{\frac{1}{2}\Sigma^{-1}\int_{0}^{t}{\beta_{s}ds}}Y_{t} - Y_{0} = \int_{0}^{t}{\sqrt{\beta_{s}}e^{\frac{1}{2}\Sigma^{-1}\int_{0}^{s}{\beta_{u}du}}dW_{s}},
\end{equation}

or writing this down in terms of $X_{t}$:

\begin{equation}
\label{sol}
X_{t} = e^{-\frac{1}{2}\Sigma^{-1}\int_{0}^{t}{\beta_{s}ds}}X_{0} + \left(I - e^{-\frac{1}{2}\Sigma^{-1}\int_{0}^{t}{\beta_{s}ds}}\right)\mu+ \int_{0}^{t}{\sqrt{\beta_{s}}e^{-\frac{1}{2}\Sigma^{-1}\int_{s}^{t}{\beta_{u}du}}dW_{s}},
\end{equation}

where $I$ is $n\times n$ identity matrix.

\subsection{Derivation of conditional distribution of $\mathbf{X_{t}}$}
\label{app:derivation_of_cond_dist}

Let $A(s) = \sqrt{\beta_{s}}e^{-\frac{1}{2}\Sigma^{-1}\int_{s}^{t}{\beta_{u}du}}$. It is a diagonal matrix and its $i$-th diagonal element $a_{ii}(s)$ equals $\sqrt{\beta_{s}}e^{-\frac{1}{2\sigma^{2}_{ii}}\int_{s}^{t}{\beta_{u}du}}$. Assume $a_{ii}(s)\in L_{2}[0,T]$ for each $i$. It\^o's integral $\int_{0}^{t}{a_{ii}(s)dW_{s}^{i}}$ is defined as the limit of integral sums when mesh of partition $\Delta$ tends to zero:

\begin{equation}
\begin{split}
    \int_{0}^{t}{a_{ii}(s)dW_{s}^{i}} = \lim_{\Delta\to 0}{\sum_{k}{a_{ii}(s_{k})\Delta W_{s_{k}}^{i}}} &\stackrel{d}{=} \lim_{\Delta\to 0}{\mathcal{N}\left(0, \sum_{k}{a_{ii}^{2}(s_{k})\Delta s_{k}}\right)} \stackrel{d}{=}\\&\stackrel{d}{=} \mathcal{N}\left(0, \lim_{\Delta\to 0}{\sum_{k}{a_{ii}^{2}(s_{k})\Delta s_{k}}}\right) = \mathcal{N}\left(0,\int_{0}^{t}{a^{2}_{ii}(s)ds}\right),
\end{split}
\end{equation}

where the first equality in distribution holds due to the properties of Brownian motion and the fact that $a_{ii}(s_{k})$ are deterministic (implying that $a_{ii}(s_{k})\Delta W_{s_{k}}^{i} = a_{ii}(s_{k})(W_{s_{k+1}}^{i} - W_{s_{k}}^{i})$ are independent normal random variables with mean $0$ and variance $a^{2}_{ii}(s_{k})(s_{k+1} - s_{k})=a^{2}_{ii}(s_{k})\Delta s_{k}$) and the second equality in distribution follows from L\'evy's continuity theorem (it is easy to check that the sequence of characteristic functions of random variables on the left-hand side converges point-wise to the characteristic function of the random variable on the right-hand side). Then, simple integration gives

\begin{equation}
    \int_{0}^{t}a_{ii}^{2}(s)ds = \int_{0}^{t}\beta_{s}e^{-\frac{1}{\sigma_{ii}^{2}}\int_{s}^{t}\beta_{u}du}ds = \int_{0}^{t}{\sigma_{ii}^{2}d\left(e^{-\frac{1}{\sigma_{ii}^{2}}\int_{s}^{t}{\beta_{u}du}}\right)} = \sigma_{ii}^{2}\left(1 - e^{-\frac{1}{\sigma_{ii}^{2}}\int_{0}^{t}{\beta_{s}ds}}\right).
\end{equation}

It implies that in multidimensional case we have:

\begin{equation}
    \int_{0}^{t}{\sqrt{\beta_{s}}e^{-\frac{1}{2}\Sigma^{-1}\int_{s}^{t}{\beta_{u}du}}dW_{s}} = \int_{0}^{t}{A(s)dW_{s}}\sim \mathcal{N}\left(0, \lambda(\Sigma, t)\right), \ \ \ \lambda(\Sigma, t) = \Sigma\left(I - e^{-\Sigma^{-1}\int_{0}^{t}{\beta_{s}ds}}\right),
\end{equation}

and it follows from (\ref{sol}) that

\begin{equation}
\label{dist}
Law(X_{t}|X_{0}) = \mathcal{N}(\rho(X_{0}, \Sigma, \mu, t), \lambda(\Sigma, t)), \ \ \ \rho(X_{0}, \Sigma, \mu, t) = e^{-\frac{1}{2}\Sigma^{-1}\int_{0}^{t}{\beta_{s}ds}}X_{0} + \left(I - e^{-\frac{1}{2}\Sigma^{-1}\int_{0}^{t}{\beta_{s}ds}}\right)\mu.
\end{equation}

\subsection{Reverse dynamics}
\label{app:reverse_dynamics}
 
The result by Anderson \yrcite{SDE-reverse} implies that if $n$-dimensional process of the diffusion type $X_{t}$ satisfies

\begin{equation}
\label{fwd}
dX_{t} = f(X_{t}, t)dt + g(t)dW_{t}, \ \ \ \ \ t\in [0,T],
\end{equation}

where $g(t)$ is a function $[0,T]\rightarrow \mathbb{R}$, then its reverse-time dynamics is given by

\begin{equation}
\label{stoch}
dX_{t} = (f(X_{t}, t) - g^{2}(t)\nabla\log{p_{t}(X_{t})})dt + g(t)d\widetilde{W}_{t}, \ \ \ t\in [0,T],
\end{equation}

where $p_{t}(\cdot)$ is the probability density function of random variable $X_{t}$ and $\widetilde{W}_{t}$ is the reverse-time standard Brownian motion such that $X_{t}$ is independent of its past increments $\widetilde{W}_{s} - \widetilde{W}_{t}$ for $s < t$. Reverse-time dynamics means that all the integrals associated with reverse-time differentials have $t$ as their lower limit (e.g. $dX_{t}$ relates to $\int_{t}^{T}{dX_{s}}=X_{T}-X_{t}$). Anderson's result is obtained under the assumption that Kolmogorov equations (for probability density functions) associated with all considered processes have unique smooth solutions. On the other hand, Song et al. \yrcite{SDE-main} argued that SDE (\ref{fwd}) has the same forward Kolmogorov equation as the following ODE:

\begin{equation}
\label{det}
dX_{t} = (f(X_{t}, t) - \frac{1}{2}g^{2}(t)\nabla\log{p_{t}(X_{t})})dt, \ \ \ t\in [0,T],
\end{equation}

which means that processes following (\ref{fwd}) and (\ref{det}) are equal in distribution if they start from the same initial distribution $Law(X_{0})$. In our case $f(X_{t}, t) = \frac{1}{2}\Sigma^{-1}(X_{t}-\mu)\beta_{t}$ and $g(t) = \sqrt{\beta_{t}}$, so we have two equivalent reverse diffusion dynamics:

\begin{equation}
    dX_{t}=\left(\frac{1}{2}\Sigma^{-1}(X_{t}-\mu)-\nabla\log{p_{t}(X_{t})}\right)\beta_{t}dt + \sqrt{\beta_{t}}d\widetilde{W}_{t}
\end{equation}

and

\begin{equation}
    dX_{t}=\frac{1}{2}\left(\Sigma^{-1}(X_{t}-\mu)-\nabla\log{p_{t}(X_{t})}\right)\beta_{t}dt,
\end{equation}

where both differential equations are to be solved backwards.

\subsection{Score estimation}
\label{app:score_estimation}

If $X_{0}$ is known, then (\ref{dist}) implies that

\begin{equation}
\begin{gathered}
    \log{p_{0t}(X_{t}|X_{0})} = -\frac{n}{2}\log{(2\pi)-\frac{1}{2}\det{\lambda(\Sigma, t)}}-\frac{1}{2}(X_{t}-\rho(X_{0}, \Sigma, \mu, t))^{\mathbf{T}}\lambda(\Sigma, t)^{-1}(X_{t}-\rho(X_{0}, \Sigma, \mu, t)) \implies
    \\
    \nabla\log{p_{0t}(X_{t}|X_{0})} = -\lambda(\Sigma, t)^{-1}(X_{t}-\rho(X_{0}, \Sigma, \mu, t)),
\end{gathered}
\end{equation}

where $p_{0t}(\cdot|X_{0})$ is the probability density function of conditional distribution $Law(X_{t}|X_{0})$. So, if we sample $X_{t}$ by the formula $X_{t}=\rho(X_{0}, \Sigma, \mu, t) + \epsilon_{t}$ where $\epsilon_{t}\sim\mathcal{N}(0,\lambda(\Sigma, t))$, then $\nabla\log{p_{0t}(X_{t}|X_{0})}=-\lambda(\Sigma, t)^{-1}\epsilon_{t}$. In the simplified case when $\Sigma=I$ we have $\lambda(I,t)=\lambda_{t}I$ where $\lambda_{t} = 1 - e^{-\int_{0}^{t}{\beta_{s}ds}}$. In this case gradient of noisy data log-density reduces to $\nabla\log{p_{0t}(X_{t}|X_{0})}=-\epsilon_{t}/\lambda_{t}$. If $\epsilon_{t}=\sqrt{\lambda_{t}}\xi_{t}$, then we have

\begin{equation}
    X_{t}=\rho(X_{0}, I, \mu, t) + \sqrt{\lambda_{t}}\xi_{t}, \ \ \ \xi_{t}\sim\mathcal{N}(0,I), \ \ \ \nabla\log{p_{0t}(X_{t}|X_{0})}=-\xi_{t}/\sqrt{\lambda_{t}}.
\end{equation}

\end{document}